\pdfoutput=1

\documentclass[11pt]{article}

\usepackage[final]{acl}

\usepackage{multirow}
\usepackage{booktabs}
\usepackage{adjustbox}
\usepackage{times}
\usepackage{latexsym}
\usepackage{amsmath}
\usepackage{makecell}
\usepackage{cleveref}
\usepackage{CJKutf8}
\usepackage{twemojis}
\usepackage{booktabs}
\usepackage{tcolorbox}
\usepackage{multirow} 
\usepackage[T1]{fontenc}
\usepackage{graphicx} 
\usepackage{float} 
\usepackage{tabularx}

\DeclareUnicodeCharacter{01DD}{\rotatebox[origin=c]{180}{e}} 
\DeclareUnicodeCharacter{0250}{\rotatebox[origin=c]{180}{a}} 
\DeclareUnicodeCharacter{1D09}{\rotatebox[origin=c]{180}{i}} 
\DeclareUnicodeCharacter{0254}{\rotatebox[origin=c]{180}{o}} 
\DeclareUnicodeCharacter{0265}{\rotatebox[origin=c]{180}{h}} 
\DeclareUnicodeCharacter{026F}{\rotatebox[origin=c]{180}{m}} 

\DeclareUnicodeCharacter{00BF}{\textquestiondown}           
\usepackage[T1]{fontenc}
\usepackage[table]{xcolor}  
\definecolor{myblue}{RGB}{230, 242, 255}  

\usepackage[utf8]{inputenc}
\usepackage{microtype}
\usepackage{graphicx}

\newcommand{\tofu}{\raisebox{-1.0ex}{\includegraphics[height=3ex]{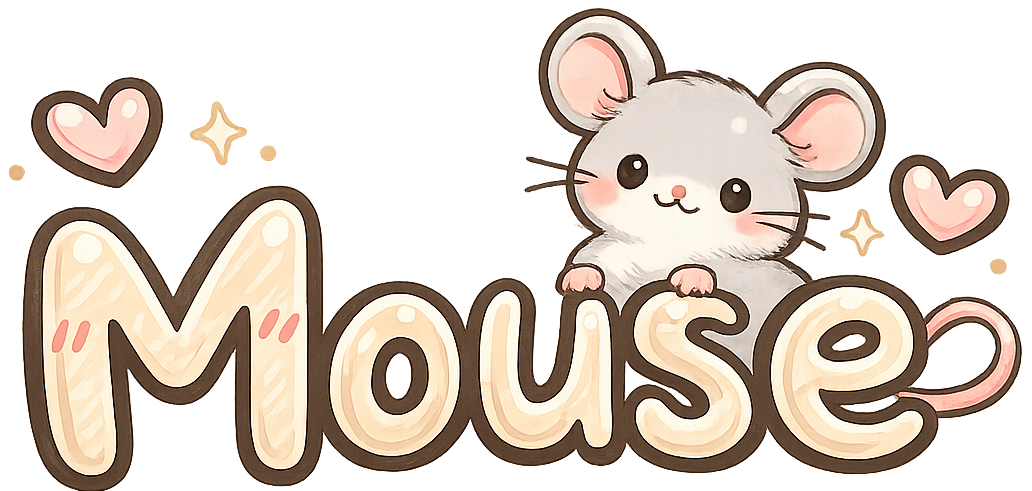}}}

\newcommand{\best}[1]{\textbf{#1}}
\newcommand{\secondbest}[1]{\underline{#1}}

\usepackage{inconsolata}
\usepackage{graphicx}
\usepackage{subcaption}
\newcommand{\zh}[1]{\begin{CJK*}{UTF8}{gbsn}#1\end{CJK*}}

\title{\tofu{} Exploring the Capability Boundaries of LLMs in Mastering of Chinese Chouxiang Language}

\author{
\textbf{Dianqing Lin}\thanks{Equal contribution},
\textbf{Tian Lan}\footnotemark[1],
\textbf{Jiali Zhu}\footnotemark[1],
\textbf{Jiang Li},
\textbf{Wei Chen},\\
\textbf{Xu Liu},
\textbf{Aruukhan},
\textbf{Xiangdong Su},
\textbf{Hongxu Hou}\thanks{Corresponding Author},
\textbf{Guanglai Gao}\\
College of Computer Science, Inner Mongolia University, China \\
\texttt{lindian7ing@163.com, velikayascarlet@gmail.com}\\
\texttt{umaru4fun@gmail.com, cshhx@imu.edu.cn}}

\usepackage{fontawesome5}

\begin{document}
\maketitle
\begin{abstract}
\textcolor{red}{\faExclamationTriangle} 
{\color{red}Warning: This paper contains content that may be offensive or harmful}

While large language models (LLMs) have achieved remarkable success in general language tasks, their performance on Chouxiang Language, a representative subcultural language in the Chinese internet context, remains largely unexplored. In this paper, we introduce Mouse, a specialized benchmark designed to evaluate the capabilities of LLMs on NLP tasks involving Chouxiang Language across six tasks. Experimental results show that, current state-of-the-art (SOTA) LLMs exhibit clear limitations on multiple tasks, while performing well on tasks that involve contextual semantic understanding. 
In addition, we further discuss the reasons behind the generally low performance of SOTA LLMs on Chouxiang Language, examine whether the LLM-as-a-judge approach adopted for translation tasks aligns with human judgments and values, and analyze the key factors that influence Chouxiang translation.
Our study aims to promote further research in the NLP community on multicultural integration and the dynamics of evolving internet languages. Our code and data are publicly available\footnote{\url{https://github.com/csdq777/Mouse}}.
\end{abstract}
\begin{figure*}[t]  
    \centering
    \includegraphics[width=0.95\textwidth]{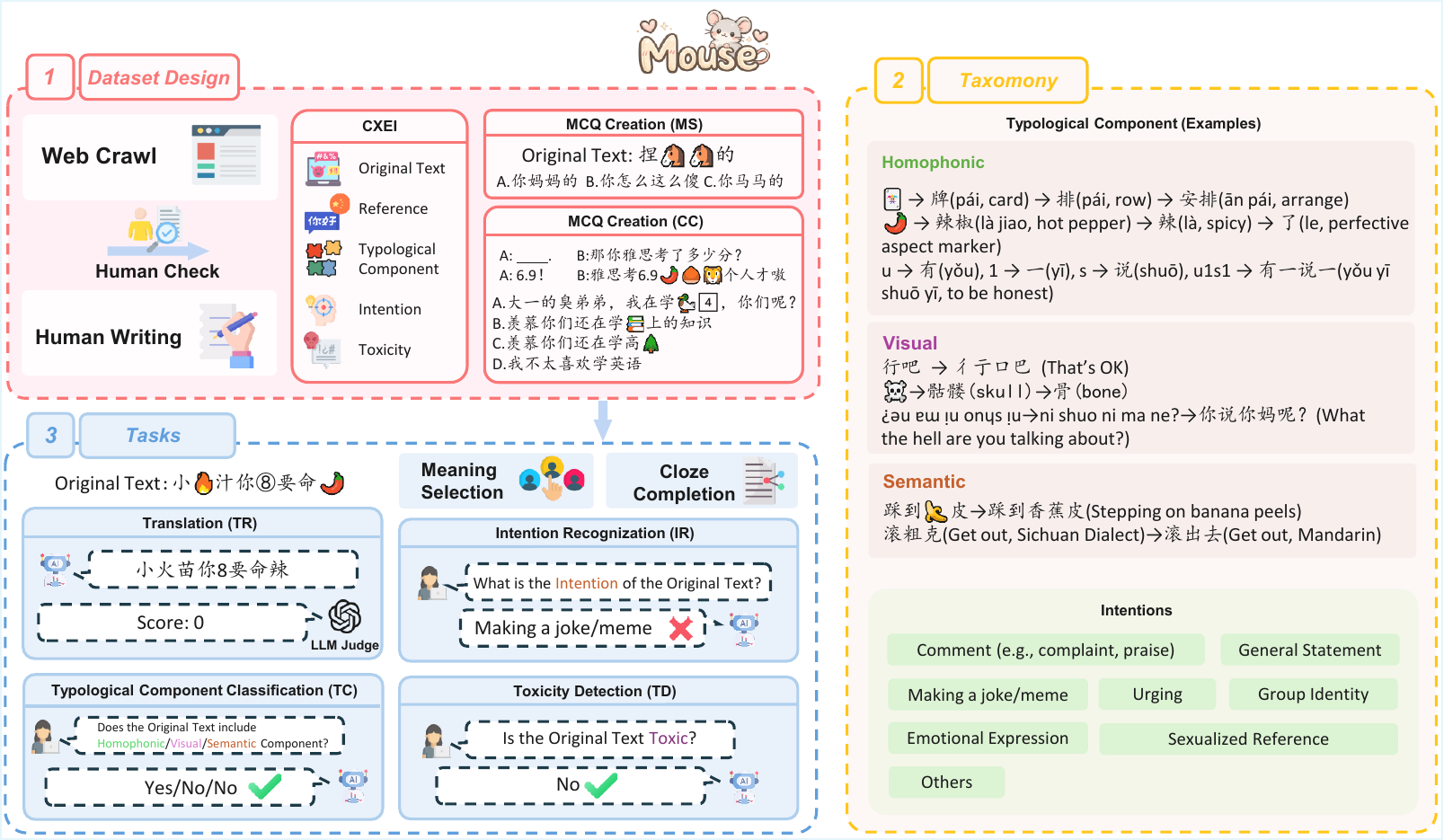} 
    \caption{Overall structure of our proposed Mouse benchmark.}
    \label{fig:overall} 
\end{figure*}
\section{Introduction}
With the widespread use of social media, internet language and memes have become an integral part of digital platforms and everyday communication~\citep{kostadinovska2018internet,vlasov2024effects}. In the Chinese internet context, Chouxiang Language represents a distinctive linguistic variant. Originating around 2015, it initially served as a mechanism to express negative sentiments and evade censorship. Consequently, the term historically carried negative connotations. However, it has evolved significantly over the past decade. Driven by the widespread popularity of Chouxiang Culture, a vast amount of non-offensive content has emerged. Chouxiang Language has thus turned into a neutral and highly symbolic subcultural code. Characterized by its specific expressive forms, it is now widely adopted by Chinese youth and online communities. A more detailed
description of the Chouxiang Culture is given in
Appendix \ref{sec:culture}.

Chouxiang Language is usually formed by transforming sentences originally composed entirely of Chinese characters into expressions that combine text, emojis, and metaphorical elements, mainly through homophonic substitution, visual symbol analogy, and literal semantic translation.
For example, in the expression "\begin{CJK}{UTF8}{gbsn}
    宁可真是个小\twemoji[height=1em]{brain} \twemoji[height=1em]{ghost}
\end{CJK}"
(\textbf{You're such a smart cookie}). The character (\begin{CJK}{UTF8}{gbsn}宁\end{CJK}) functions as a homophonic substitute for "You" (\begin{CJK}{UTF8}{gbsn}你\end{CJK}); the emoji "{\twemoji[height=1em]{brain}}" metaphorically implies "cleverness" through the visual association of a brain; and "\twemoji[height=1em]{ghost}" retains the literal semantics of "ghost(\begin{CJK}{UTF8}{gbsn}鬼\end{CJK})." Although this mode of expression significantly deviates from Standard Chinese in both form and semantics, thereby creating a non-standard semantic space, it maintains high intelligibility within communities that share the same subcultural context.

Despite the widespread influence of Chouxiang Language as a representative internet language within the Chinese internet and society, a systematic analysis of this phenomenon remains absent in the existing natural language processing (NLP) community. Particularly, given the remarkable performance of Large Language Models (LLMs) across various NLP tasks in recent years~\citep{NEURIPS2020_1457c0d6,achiam2023gpt,liu2024deepseek}, a interesting question arises: What are the capabilities of LLMs in mastering Chouxiang Language?

We consider this problem important for three reasons: First, from the perspective of computational social science and culture, existing LLMs and benchmarks exhibit a pronounced Western-centric bias, predominantly reflecting Western mainstream values~\citep{cao-etal-2023-assessing,naous-etal-2024-beer,durmus2024towards,singh-etal-2025-global}. Since language is the carrier of the cultural core~\citep{wang-etal-2024-best,zhang-etal-2024-mc2,wang-etal-2025-multilingual,wang-etal-2026-chatgpt}, exploring Chouxiang Language, a typical non-Western subcultural linguistic variant, is essential. It not only fills the gap in multicultural research for LLMs but is also crucial for understanding linguistic practices within complex cultural contexts. 
Second, existing studies focusing on Chinese internet language often confine such linguistic phenomena to negative pragmatic dimensions, such as toxic language detection and perturbed language detection~\citep{xiao-etal-2024-toxicloakcn,wu-etal-2025-enhancing-chinese,bai-etal-2025-state,guo-etal-2025-lost}. This focus overlooks the neutral and even positive functions that have emerged during the long-term evolution of Chouxiang Language. These non-negative semantic spaces remain largely underexplored. Finally, although prior studies have made impressive progress in the study of Chinese memes and Chinese buzzwords~\citep{xie-etal-2025-large,huang-etal-2025-large}, these are only a subset of Chouxiang Language. Given that Chouxiang Language possesses more complex semantic structures and linguistic features, this paper aims to bridge this research gap. We strive to construct a more comprehensive analytical framework of Chouxiang Language for the NLP community, thereby fostering a deeper understanding of such online linguistic phenomena.

To bridge this gap, we introduce Mouse, a benchmark designed to evaluate LLMs' proficiency in Chouxiang Language across six tasks. Our results show that while these LLMs demonstrate some understanding of contextual information, they have difficulty handling other aspects. In addition, we conducted a detailed analysis, hoping that our study can contribute to the development of the NLP community focused on subcultural languages.

In summary, the main contributions of this paper are as follows:

\begin{itemize}
    \item \textbf{Subculture Formalization}~We introduce Chouxiang Language, a unique internet subcultural language, to the NLP community. 
    \item \textbf{Evaluation Benchmark}~We propose Mouse, the first LLM evaluation benchmark tailored for Chouxiang Language. Comprising six NLP tasks, aiming to evaluate LLMs' processing of this subcultural language.
    \item\textbf{Experimental Analysis}~We conduct extensive experiments on SOTA LLMs. Furthermore, we analyze the potential factors underlying their performance and offer insights for future research.
\end{itemize}

\section{Preliminaries}
\begin{CJK}{UTF8}{gbsn}
\begin{table*}[t]
\centering
\small
\resizebox{\textwidth}{!}{
\begin{tabular}{lcccp{3cm}}
\toprule
\textbf{Component} & \textbf{Original Text} & \textbf{Derivational Logic} & \textbf{Standard Chinese} & \textbf{English Reference} \\ \midrule
\multirow{2}{*}{\textbf{Homophonic}} & 主包 (zhǔ bāo) & Near-homophone substitution & 主播 (zhǔ bō) & Streamer \\
 & 91安\twemoji{joker}上 & \twemoji{joker} $\rightarrow$ 牌 (pái) $\rightarrow$ 排 (pái) & 91安排上 & Arrange you with 91 \\ \midrule
\multirow{3}{*}{\textbf{Visual}} & 彳亍口巴 & Structural decomposition of characters & 行吧 (xíng ba) & That's OK \\
 & 我扬了你\twemoji{skull}灰 & Iconographic metaphor (\twemoji{skull} $\rightarrow$ 骨) & 我扬了你骨灰 & Scatter your ashes \\
 & ¿ǝu ɐɯ ᴉu onɥs ᴉu & Inverted Pinyin & 你说你妈呢 & What the hell are you talking about \\ \midrule
\multirow{2}{*}{\textbf{Semantic}} & 踩到\twemoji{banana}皮 & Direct symbolic literalism & 踩到香蕉皮 & Step on a banana peel \\
 & 滚粗克 (gǔn cū kè) & Dialectal register transformation & 滚出去 (gǔn chū qù) & Get out \\ \bottomrule
\end{tabular}}
\caption{Representative examples across three representational components of Chouxiang Language.}
\label{tab:component-cases}
\end{table*}

\end{CJK}
\subsection{The Definition of Chouxiang Language}
Chouxiang Language is a distinctive variant of Chinese internet language. It serves as a concrete manifestation of Chinese online subculture. Its core mechanism integrates diverse elements, such as special characters, homophones, Pinyin acronyms, dialects, emojis, Chinese radical combinations, and internet memes~\citep{chen2021chouxiang}. Characterized by its implicit nature where meanings are felt rather than explicitly stated, Chouxiang Language functions as a subcultural mode of communication that emphasizes the conveyance of emotion over literal information.
\subsection{Taxonomy}
To systematically analyze the complexity of Chouxiang Language and clarify its underlying logic, we categorize it into two dimensions: representational components and intents. This fine-grained taxonomy provides the theoretical foundation for our subsequent evaluation. By jointly modeling linguistic structure and pragmatic function, the taxonomy enables a more comprehensive evaluation of model capabilities.

\subsubsection{The Representational Component of Chouxiang Language}
Prior studies~\citep{chen2021chouxiang} primarily categorized Chouxiang Language based on its origins, dividing it into symbols, homophones, dialects, and memes. Although these classifications documented early linguistic phenomena, they exhibit significant feature overlap and fail to capture recent, more deconstructive practices. Consequently, we propose a systematic classification of representational components from the perspective of symbolic representation~\citep{shelestiuk2003semantics}. We categorize these components into three core dimensions: homophonic, visual, and semantic. Within this framework, a single sentence may simultaneously exhibit characteristics from multiple dimensions. The examples across three representational components can be found in Table~\ref{tab:component-cases}.

\paragraph{Homophonic Component} This dimension exploits the phonological redundancy of the Chinese language. Users construct Chouxiang expressions through homophonic substitution using Chinese characters, alphanumeric symbols, or multi-stage “image–name–homophone” mapping chains. This process maps the target vocabulary to characters with similar or identical pronunciations.

\paragraph{Visual Component}
Leveraging the ideographic nature of Chinese characters and the pictographic properties of emojis, this component exploits visual analogy through geometric structures, radicals, and other iconic imagery and emoji. It manifests through three mechanisms:
(1) Character Decomposition, which fragments glyphs into constituent radicals to increase textual discreteness;
(2) Visual Metaphor, where characters and emojis undergo semantic extension based on intuitive visual associations;
and (3) Geometric Transformation, involving inverted or deformed typography to disguise sensitive content.

\paragraph{Semantic Component} This dimension focuses on meaning-level mapping. It includes (1) Symbolic Literalism, which uses the direct or socially shared meanings of emojis, and (2) Dialectal Borrowing, which draws on regional pronunciation or writing variants to add humor or shift style while preserving the core meaning.

\subsubsection{The Intent of Chouxiang Language}
In contemporary social media, Chouxiang Language serves not merely as a marker of identity but also functions as a vehicle for diverse communicative intents, akin to natural language. These communicative acts include, but are not limited to: comments of specific events (e.g., sarcasm or praise), direct emotional expressions (e.g., venting anger or helplessness), basic factual statements, and subculturally characteristic humor and memes. Furthermore, within specific contexts, it exhibits action-oriented directives or functions as a tool for implicit sexual reference. 

As Chouxiang Language enters broader use, analysis must move beyond surface-level symbols and consider its role in social interaction and behavioral intent. Consequently, we categorize these intents into eight distinct classes: Comment (e.g., complaint, praise), Emotional Expression, General Statement, Sexualized Reference, Making a Joke \& Memes, Group Identity, Urging, and Others.

\zh{

\begin{table*}[ht]
\centering
\small 
\begin{tabular}{lll}
\toprule
\textbf{Attribute} & \textbf{Example (ZH)} & \textbf{Example (EN)} \\ 
\midrule
Original Text & 小\twemoji[height=1em]{fire}汁你⑧要命\twemoji[height=1em]{hot pepper}  & N/A \\
Reference & 小伙子你不要命了 & Young man, are you out of your mind? \\
Representational Component& 谐音 & Homophonic \\
Intent & 评价（吐槽，赞扬等） & Comments (criticisms, praises, etc.) \\
Toxicity & 0 & 0 \\
\bottomrule
\end{tabular}
\caption{Chinese and English examples for each attribute in a CXEI. The conversion process is as follows: \twemoji{fire} $\rightarrow$ 火 (huǒ) $\rightarrow$ 伙 (huǒ); 汁 (zhī) $\rightarrow$ 子 (zi); ⑧$\rightarrow$八 (bā) $\rightarrow$ 不 (bù); and \twemoji[height=0.9em]{hot pepper} $\rightarrow$ 辣椒 (là jiāo) $\rightarrow$ 辣 (là) $\rightarrow$ 了 (le).}
\label{tab:chouxiang-example}
\end{table*}

\subsection{Chouxiang Language Evaluation Instance}
Drawing inspiration from McBE~\citep{lan-etal-2025-mcbe}, we integrate the Chouxiang Language Evaluation Instance (CXEI) into Mouse, which is a structured evaluation concept. As the core unit of our benchmark, CXEI enables a detailed assessment of model performance in processing Chouxiang Language.
}
Mouse comprises a total of 1,099 CXEIs. Each CXEI is characterized by the following attributes:

\paragraph{Original Text} The raw text in Chouxiang Language, typically composed of a mixture of emojis, Chinese characters, Latin letters, punctuation marks and other characters.

\paragraph{Reference} The corresponding text consisting exclusively of Chinese, serving as an translation.

\paragraph{Representational Component} These are categorized into three types: Homophonic, Visual, and Semantic.

\paragraph{Intent} The categories include Comment (e.g., Complaint, Praise), Emotional Expression, General Statement, Sexualized Reference, Humor \& Memes, Urging, Group Identity, and Others.

\paragraph{Toxicity} A binary label indicating whether the text contains toxic content (labeled as 1 for toxic, and 0 otherwise).

An example of CXEI can be found in Table~\ref{tab:chouxiang-example}.

\section{Dataset and Evaluation Tasks}
The data sources for Mouse fall into two primary categories: web collection and manual construction. We aim to harvest data from diverse contexts to ensure the breadth, depth, and representativeness of the dataset. More detailed are provided in the Appendix \ref{sec:dataset}.

\paragraph{Task Definition} Systematically quantifying LLM proficiency in Chouxiang Language and culture presents significant challenges. We address this by introducing six tasks in Mouse: Translation, Representational Component Classification, Intent Recognition, Toxicity Detection, Meaning Selection, and Cloze Completion. Higher scores on these metrics indicate better proficiency of meaning. Detailed evaluation prompts are provided in the Appendix~\ref{sec:Prompt}.

\subsection{Translation (TR)}
The Translation task aims to rigorously assess a model's ability to decode complex Chouxiang Language and translate it into Standard Chinese. In this task, models are required to accurately identify and reconstruct pragmatic information encoded in original text. The ultimate goal is to generate coherent and accurate sentences in Standard Chinese.


We evaluate model performance using an LLM-as-a-Judge~\citep{NEURIPS2023_91f18a12}. To ensure evaluation integrity, an anti-cheating filter is first applied, where responses that merely replicate the original text without meaningful transformation are automatically assigned a score of 0. For valid responses, we score them on three levels based on how closely their meaning aligns with human references: 2 points are awarded for full consistency in both core semantics and key terms; 1 point is given for partial alignment where essential information is retained despite minor deviations; and 0 points are assigned for complete divergence or a total loss of key information.
To obtain the final task score, we average the results of all CXEIs and map them linearly to a 0–1 scale.







\subsection{Representational Component Classification (RC)}
The Representational Component Classification task aims to evaluate whether a given original text contains specific representative components, such as semantic, homophonic, or visual features. The task is conceptually designed as three parallel binary classification sub-tasks. However, considering the models’ understanding capability, we separate them into three independent tasks and then aggregate the results. For each sub-task, the model independently determines whether the given component is present in the text (1 for presence, 0 for absence). The final score is obtained by computing the Balanced Accuracy for each sub-task and averaging the three results.


\subsection{Intent Recognition (IR)}
The Intent Recognition task measures the model's ability to identify the underlying purpose of Chouxiang Language expressions. Models must classify original texts into predefined categories that reflect the subculture's pragmatic functions: informational (General Statement, Urging), affective (Comment, Emotional Expression), social (Group Identity, Humor \& Memes), and non-standard semantic deviations (Sexualized Reference). Performance is measured by classification accuracy, defined as the proportion of correctly identified intents.

\subsection{Toxicity Detection (TD)}
The Toxicity Detection task evaluates a model's ability to identify veiled malicious speech that bypasses traditional keyword filters through emojis, acronyms, or homophones. Specifically, models must distinguish genuine aggression consisting of verbal abuse, hate speech, and extreme sarcasm from benign interactions in subcultural context such as self-deprecation or irony. Framed as a binary classification, this task requires the model to output a single integer (0 or 1) to indicate the presence of toxicity.

\subsection{Meaning Selection (MS)}
The Meaning Selection task evaluates semantic precision by requiring models to identify the correct meaning of a original text from multiple candidates. Conducted without conversational context, this task focuses on the model’s core semantic understanding and its ability to catch target meaning. Performance is measured by accuracy, representing the proportion of correct selections.

\subsection{Cloze Completion (CC)}
The Cloze Completion task aims to assess the model's capability to accurately employ Chouxiang Language within specific social contexts. In contrast to the Meaning Selection task, this task requires the model to select the most natural and contextually appropriate option from a set of candidates, based on the provided conversational logic and emotional tone. Consequently, this task prioritizes the evaluation of the model's contextual adaptability and deep semantic alignment. Similar to the previous tasks, performance is measured using accuracy.

\section{Evaluting Chouxiang Language in LLMs}
\begin{table*}[t] 
    \centering
    \small 
    \setlength{\tabcolsep}{4pt} 
    \begin{tabular*}{\textwidth}{@{\extracolsep{\fill}}l cc cc cc cc c c@{}}
        \toprule
        \multirow{2}{*}{\textbf{Model}} 
& \multicolumn{2}{c}{\textbf{TR}} 
& \multicolumn{2}{c}{\textbf{RC (Bal.)}} 
& \multicolumn{2}{c}{\textbf{IR (Bal.)}} 
& \multicolumn{2}{c}{\textbf{TD}} 
& \multicolumn{1}{c}{\textbf{MS}} 
& \multicolumn{1}{c}{\textbf{CC}}\\
        
        \cmidrule(lr){2-3} \cmidrule(lr){4-5} \cmidrule(lr){6-7} \cmidrule(lr){8-9} \cmidrule(lr){10-10} \cmidrule(lr){11-11} 
        & Sem.Acc. & PR. & Bal.Acc. & Macro-F1 & Bal.Acc. & Macro-F1 & Acc. & F1 & Acc. & Acc. \\
        \midrule
        DeepSeek-V3.2 & \best{0.494} & \best{0.449} & 0.533 & 0.361 & 0.255 & 0.245 & \secondbest{0.731} & \best{0.728} & 0.750 & 0.530 \\
        Doubao-Seed-1.8 & \secondbest{0.448} & \secondbest{0.408} & \secondbest{0.551} & \best{0.542} & \best{0.317} & \best{0.275} & \best{0.751} & 0.687 & \best{0.940} & \best{0.780} \\
        Qwen3-Max & 0.404 & 0.348 & 0.538 & 0.367 & 0.263 & 0.237 & 0.655 & 0.683 & \secondbest{0.830} & \secondbest{0.670} \\
        GPT5.2 & 0.441 & 0.397 & \best{0.562} & 0.445 & \secondbest{0.280} & \secondbest{0.256} & 0.694 & \secondbest{0.698} & 0.820 & 0.560 \\
        Mistral-Large-3 & 0.227 & 0.199 & 0.541 & 0.438 & 0.201 & 0.185 & 0.622 & 0.663 & 0.630 & 0.500 \\
        \midrule
        Qwen3-32B & 0.209 & 0.168 & 0.510 & 0.295 & 0.189 & 0.141 & 0.605 & 0.637 & 0.660 & 0.490 \\
        Qwen3-14B & 0.228 & 0.168 & 0.518 & 0.324 & 0.173 & 0.132 & 0.656 & 0.626 & 0.560 & 0.400 \\
        Qwen3-8B & 0.168 & 0.126 & 0.523 & 0.336 & 0.145 & 0.075 & 0.644 & 0.636 & 0.520 & 0.370 \\
        Qwen3-4B & 0.144 & 0.099 & 0.501 & 0.273 & 0.126 & 0.038 & 0.470 & 0.632 & 0.490 & 0.340 \\
        Qwen3-1.7B & 0.096 & 0.066 & 0.500 & 0.272 & 0.136 & 0.069 & 0.573 & 0.225 & 0.270 & 0.350 \\
        Qwen3-0.6B & 0.041 & 0.018 & 0.500 & 0.272 & 0.125 & 0.050 & 0.520 & 0.487 & 0.180 & 0.280 \\
        \midrule
        Ministral-3-14B & 0.117 & 0.091 & 0.518 & \secondbest{0.476} & 0.188 & 0.170 & 0.541 & 0.645 & 0.480 & 0.340 \\
        Ministral-3-8B & 0.093 & 0.076 & 0.500 & 0.415 & 0.157 & 0.108 & 0.540 & 0.647 & 0.380 & 0.330 \\
        Ministral-3-3B & 0.053 & 0.046 & 0.504 & 0.380 & 0.180 & 0.138 & 0.567 & 0.625 & 0.250 & 0.230 \\
        \bottomrule
    \end{tabular*}
    \vspace{2mm}
    \caption{Main results of model evaluation on Chouxiang Language across NLP tasks. Metrics: \textbf{Sem.Acc.} Semantic Accuracy represents the percentage of the score attained; \textbf{PR.} Perfect Rate is the proportion of perfectly reconstructed translations; \textbf{Bal.Acc.} denotes Balanced Accuracy. \best{Bold}: best; \secondbest{underline}: second-best.}
    \label{tab:model_main}
\end{table*}
\subsection{Experimental Setup}
\paragraph{Model}In our experiments, we evaluated two distinct groups of models. The first group consists of locally deployed LLMs with relatively small parameter sizes, including the Qwen3 Dense family (0.6B, 1.7B, 4B, 8B, 32B)~\citep{yang2025qwen3} and the Mistral family (3B, 8B, 14B)~\citep{mistral2025introducing}. For open-source and commercial models exceeding 32B parameters, we utilized API-based access to ensure feasibility within our budget constraints. These models include Qwen3Max~\citep{yang2025qwen3}, GPT-5.2~\citep{openai2025gpt52systemcard}, DeepSeek-V3.2~\citep{liu2024deepseek}, Doubao-Seed-1.8~\citep{seedseed1}, and Mistral-3-Large~\citep{mistral2025introducing}.

For all locally deployed models were evaluated on two RTX PRO 6000 (96GB) GPUs. During inference, we set the temperature to 0 and the maximum output length to 128 tokens, while keeping all other hyperparameters at their default values. 

\paragraph{Metrics}For our evaluation, we employ multiple metrics to ensure a robust assessment. For the translation task, automatic metrics such as BLEU~\citep{post-2018-call}, chrF++~\citep{popovic-2017-chrf}, and COMET~\cite{rei-etal-2020-comet} are not well suited to our setting, because Chouxiang Language remains largely grounded in Chinese, with other “Chouxiang” elements mainly serving as substitutions for specific words. If directly used to compare Chouxiang sentences with target sentences, these metrics may produce artificially high scores due to substantial surface-level overlap in Chinese expressions. As a result, they cannot accurately reflect the model’s actual ability to understand and interpret Chouxiang Language. Therefore, for the translation task, we adopt a rule-based LLM-as-a-Judge using DeepSeek-V3.2. All other metrics are implemented using the scikit-learn library \citep{scikit-learn}. Specifically, we use Accuracy and F1 score for the Toxicity Detection task where the data is balanced. For Meaning Selection and Cloze Completion, we also report Accuracy. For other tasks involving imbalanced data, such as Representational Component Classification and Intent Recognition, we adopt Balanced Accuracy and Macro-F1 to better measure the effectiveness of Mouse, as metric selection can significantly influence system rankings in imbalanced contexts. It should be noted that in the RC task, Bal.Acc. is calculated by averaging the Balanced Accuracy of the phonetic, visual, and semantic categories. Furthermore, we apply the Matthews Correlation Coefficient (MCC)~\cite{matthews1975comparison} to investigate potential model hallucinations in classification. To assess the alignment between LLM-as-a-judge and human judgment, we utilize Quadratic Weighted Kappa (QWK)~\cite{article-kappa} as the primary evaluation metric, which provides a standardized measure of agreement beyond chance.

\subsection{Results}
\begin{table}[t] 
    \centering
    \resizebox{\columnwidth}{!}{
    \normalsize
    \setlength{\tabcolsep}{6pt} 
    \begin{tabular}{l ccc cc cc} 
        \toprule
        \multirow{2}{*}{\textbf{Model}} & \multicolumn{3}{c}{\textbf{RC}} & \multirow{2}{*}{\textbf{IR}} & \multirow{2}{*}{\textbf{TD}} & \multirow{2}{*}{\textbf{MS}} & \multirow{2}{*}{\textbf{CC}} \\
        \cmidrule(lr){2-4} 
        & Homophonic & Semantic & Visual & & & & \\
        \midrule
        DeepSeek-V3.2   & 0.118 & 0.031 & \secondbest{0.110} & 0.186 & \secondbest{0.471} & 0.625 & 0.393 \\
        Doubao-Seed-1.8 & \secondbest{0.191} & 0.076 & 0.060 & \best{0.218} & \best{0.500} & \best{0.912} & \best{0.692} \\
        Qwen3-Max       & 0.179 & 0.038 & \secondbest{0.110} & 0.179 & 0.349 & \secondbest{0.751} & \secondbest{0.548} \\
        GPT5.2          & \best{0.192} & 0.044 & \best{0.164} & \secondbest{0.191} & 0.404 & 0.733 & 0.401 \\
        Mistral-Large-3 & 0.061 & \secondbest{0.079} & 0.104 & 0.116 & 0.292 & 0.447 & \secondbest{0.360} \\
        \midrule
        Qwen3-32B       & 0.033 & 0.040 & 0.083 & 0.108 & 0.243 & 0.491 & 0.324 \\
        Qwen3-14B       & 0.068 & 0.015 & 0.093 & 0.115 & 0.308 & 0.342 & 0.219 \\
        Qwen3-8B        & 0.022 & \best{0.083} & 0.104 & 0.048 & 0.294 & 0.281 & 0.185 \\
        Qwen3-4B        & 0.000 & 0.000 & 0.020 & 0.014 & 0.107 & 0.237 & 0.153 \\
        Qwen3-1.7B      & 0.000 & 0.000 & 0.000 & 0.022 & 0.126 & -0.098 & 0.181 \\
        Qwen3-0.6B      & 0.000 & 0.000 & 0.000 & 0.000 & 0.036 & -0.233 & 0.164 \\
        \midrule
        Ministral-3-14B & 0.015 & 0.063 & 0.028 & 0.079 & 0.194 & 0.224 & 0.202 \\
        Ministral-3-8B  & 0.022 & -0.031 & 0.020 & 0.050 & 0.200 & 0.072 & 0.160 \\
        Ministral-3-3B  & -0.052 & 0.038 & 0.053 & 0.071 & 0.186 & -0.127 & 0.118 \\
        \bottomrule
    \end{tabular}}
    \caption{MCC Results. This table measures the correlation between LLM predictions and ground truth labels across various tasks. \best{Bold}: best; \secondbest{underline}: second-best.}
    \label{tab:model_mcc}
\end{table}

We report the performance of six tasks on fourteen LLMs for Chouxiang Language in Table~\ref{tab:model_main}. The results of human performance can be found in Appendix~\ref{sec:human-performance}.

\paragraph{Model Scale} Evaluation of the Qwen and Mistral series (including Qwen3-Max and Mistral-3-Large) confirms that scaling generally improves performance, which is consistent with previous findings~\citep{xuan-etal-2025-mmlu}. However, Qwen3 shows a counterintuitive decline in performance from 14B to 32B on most tasks. This suggests that, compared with the 14B model, scaling up to 32B may trigger an “overthinking” effect without bringing a qualitative improvement in performance on NLP tasks. Appendix \ref{sec:appendix_case_study} provides detailed experimental validation.

\begin{figure}[t]
    \centering

    \includegraphics[width=\columnwidth]{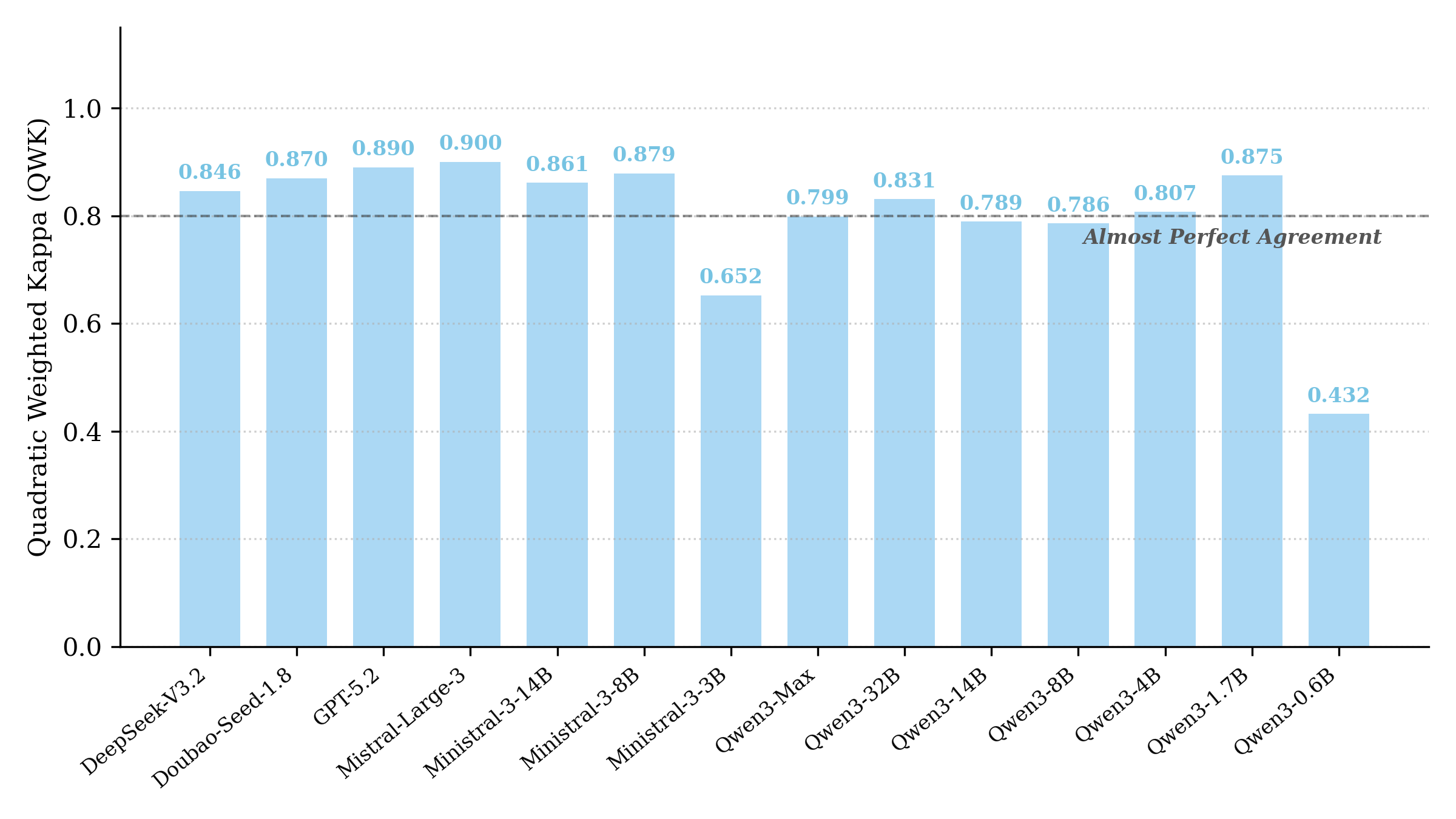} 
    
    \caption{Inter-rater reliability of the LLM-as-judge.}
    \label{fig:validation_qwk}
\end{figure}

\paragraph{Inconsistent Performance Across NLP Tasks} By comparing various commercial closed-source models, we find that the performance of SOTA LLMs is highly competitive, with Doubao-Seed-1.8 leading across multiple NLP tasks. Our study reveals that all LLMs still face significant challenges in zero-shot tasks (Translation, Representational Component, Intent Recognition) involving Chouxiang language. Performance is acceptable in Toxicity Detection and Cloze Completion, and it is excellent in Meaning Selection; however, a notable performance gap remains. Specifically, despite both being zero-shot binary classification tasks, the accuracy of Toxicity Detection  is approximately 15\% higher than that of Representational Component. This indicates that for current state-of-the-art LLMs, detecting toxic language in Chouxiang language is much easier than identifying its representational components. This disparity likely stems from a historical research bias toward the toxic analysis of Chouxiang language~\citep{xiao-etal-2024-toxicloakcn,wu-etal-2025-enhancing-chinese}, which has made models more sensitive to toxic features. Consequently, existing research has largely overlooked the functional role of Chouxiang language as a means of social interaction, resulting in poor model performance on Representational Component tasks that require understanding the logic of linguistic transformation.

\begin{figure}[t]
    \centering

    \includegraphics[width=\columnwidth]{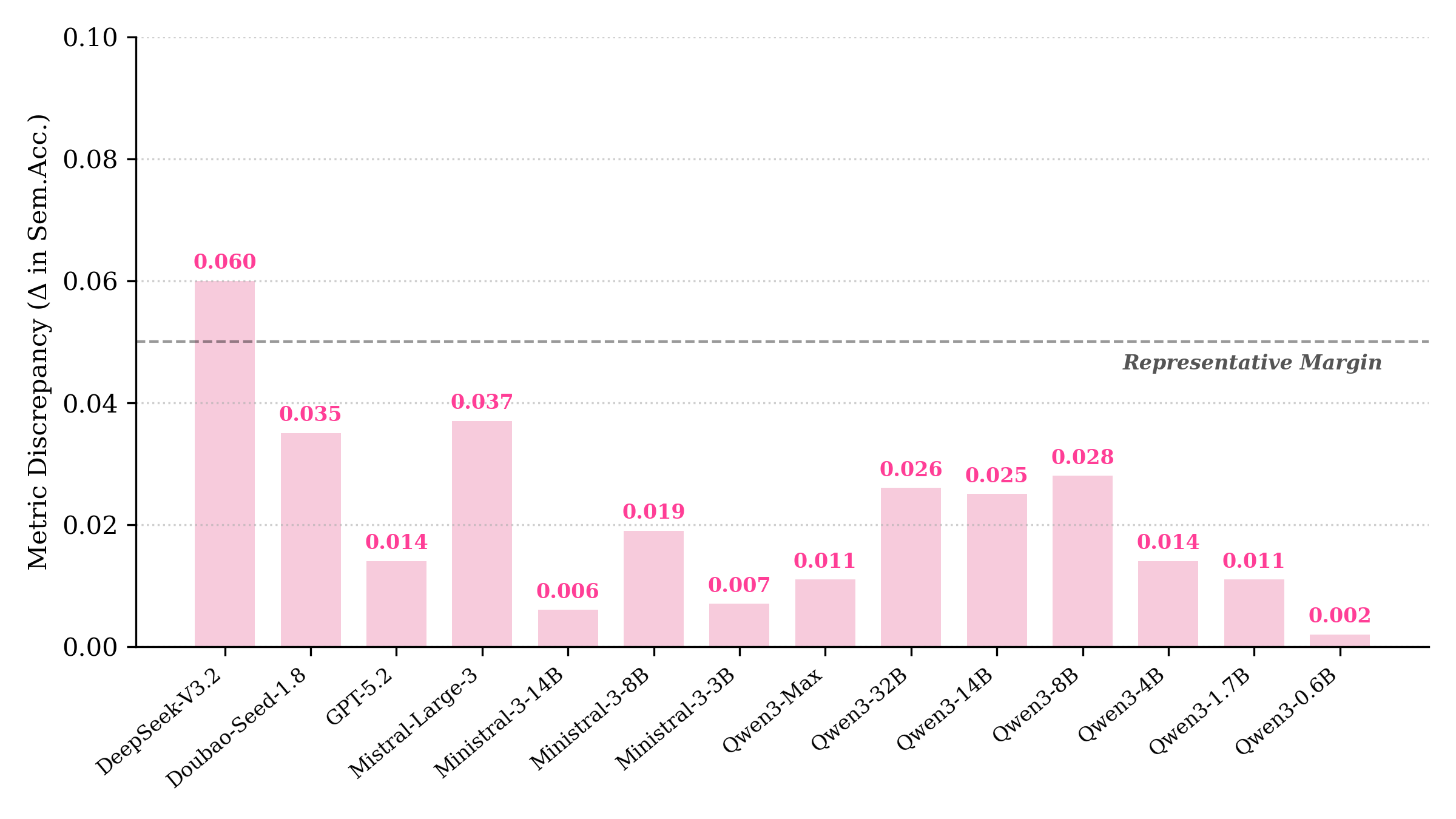} 
    \caption{Statistical stability of the qualitative sample.}
    \label{fig:validation_delta}
\end{figure}

\section{Discussion}

\subsection{Do LLMs Exhibit Hallucinations in Classification Tasks?}

The reliability of model evaluation on Chouxiang Language tasks can be effectively quantified through the MCC. Unlike simple accuracy, MCC provides a robust measure of the correlation between predicted and actual classifications. As shown in Table~\ref{tab:model_main} and Table~\ref{tab:model_mcc}, we observe a strong positive correlation between overall task performance and MCC values. Commercial closed-source models consistently maintain higher MCC values, reflecting a stable alignment across all evaluated NLP tasks.

Conversely, the performance of small-scale or open-source models often collapses in high-complexity tasks such as Representational Component. In these instances, MCC values frequently drop toward zero or even into negative territory. An MCC near zero suggests that the model is performing at a level equivalent to random chance, indicating that its outputs are driven by stochastic guessing rather than genuine pragmatic inference. Negative MCC values represent a more severe form of hallucination, where the model exhibits a systemic misinterpretation of the camouflaged signal, consistently assigning incorrect labels based on misleading surface-level patterns.

Furthermore, individual models often display wide discrepancies in MCC across different tasks, reflecting varying levels of task difficulty. When comparing Toxicity Detection, Meaning Selection, and Cloze Completion against the more challenging Representational Component and Intent Recognition tasks, models generally yield significantly lower MCC values on the latter. Notably, even on seemingly straightforward tasks like Meaning Selection, a profound capability gap persists. While DouBao-Seed-1.8 can achieve an MCC as high as 0.912, small-scale models still exhibit negative MCC values. 

\subsection{How Does the Performance of LLM-as-judge Compare to that of Humans?}


To ensure the reliability of the automated evaluation, we adopt a two-stage validation procedure that assesses both inter-rater agreement and statistical stability. First, a human correlation study conducted on a qualitative subset ($N=150$) examines the agreement between expert human annotations and the LLM-as-judge. The resulting QWK scores typically range from 0.8 to 0.9. As illustrated in Figure~\ref{fig:validation_qwk}, these values fall within the ``Almost Perfect Agreement'' range ($QWK \ge 0.8$), indicating a high level of consistency between the automated judge and human judge.

Second, to assess the generalizability of this subset, we compare the sampled CXEI original–reference pairs ($N=150$) with the full set of CXEI original–reference pairs ($N=1099$). As shown in Figure~\ref{fig:validation_delta}, the absolute difference in Semantic Accuracy generally remains within a margin of 0.05, suggesting that the sampled subset is broadly representative of the full dataset. Together, these results indicate that the LLM-as-a-judge approach provides a reliable and statistically stable evaluation criterion for assessing LLM performance on Chouxiang language translation task.

\subsection{What Factors Determine the Quality of Chouxiang Language Translation?}
\begin{figure}[t]
    \centering

    \includegraphics[width=\columnwidth]{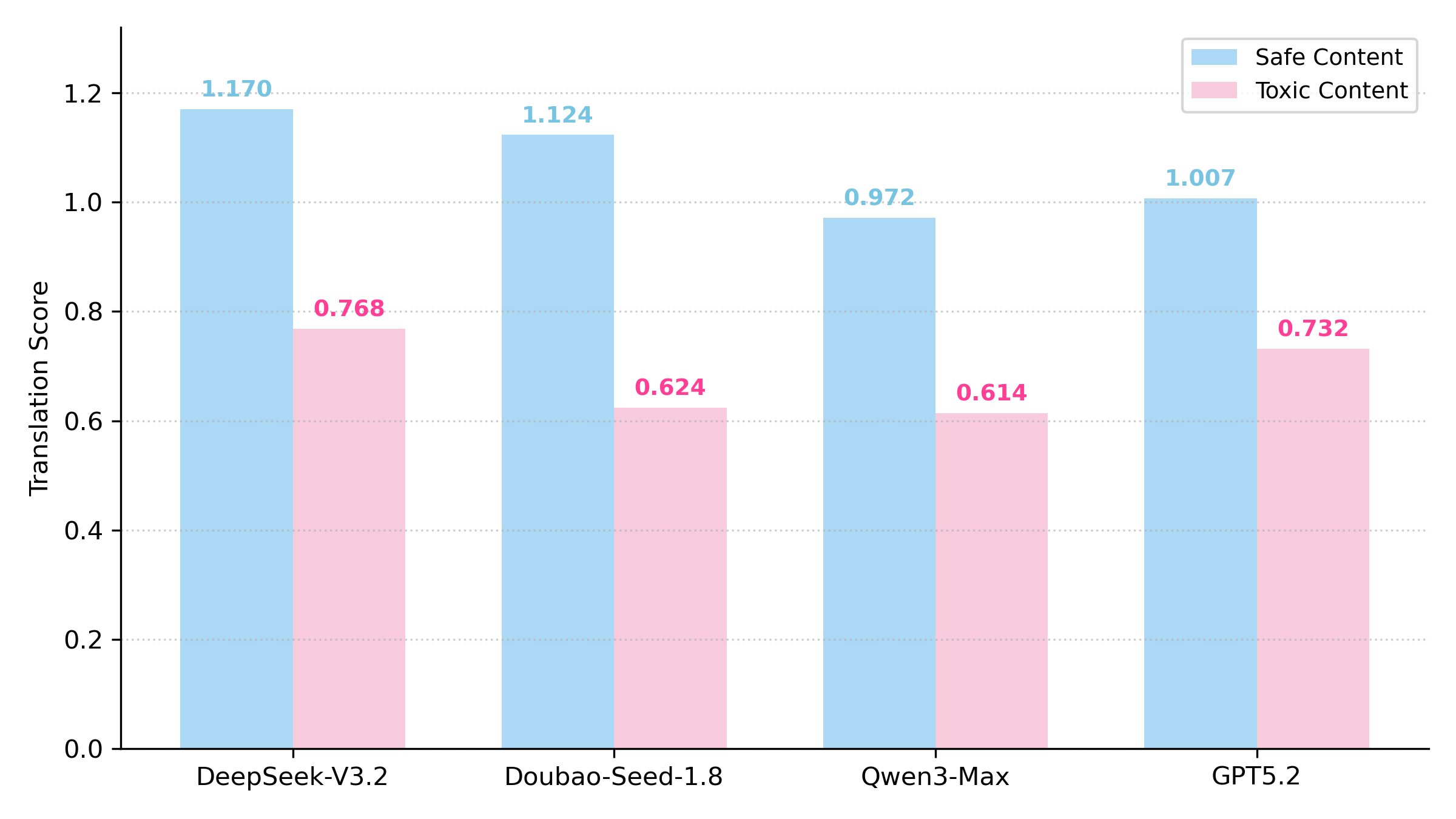} 
    \caption{Impact of toxic contexts on LLM translation Fidelity.}
    \label{fig:translation_gap_commercial_open}
\end{figure}

\begin{figure}[t]
    \centering

    \includegraphics[width=\columnwidth]{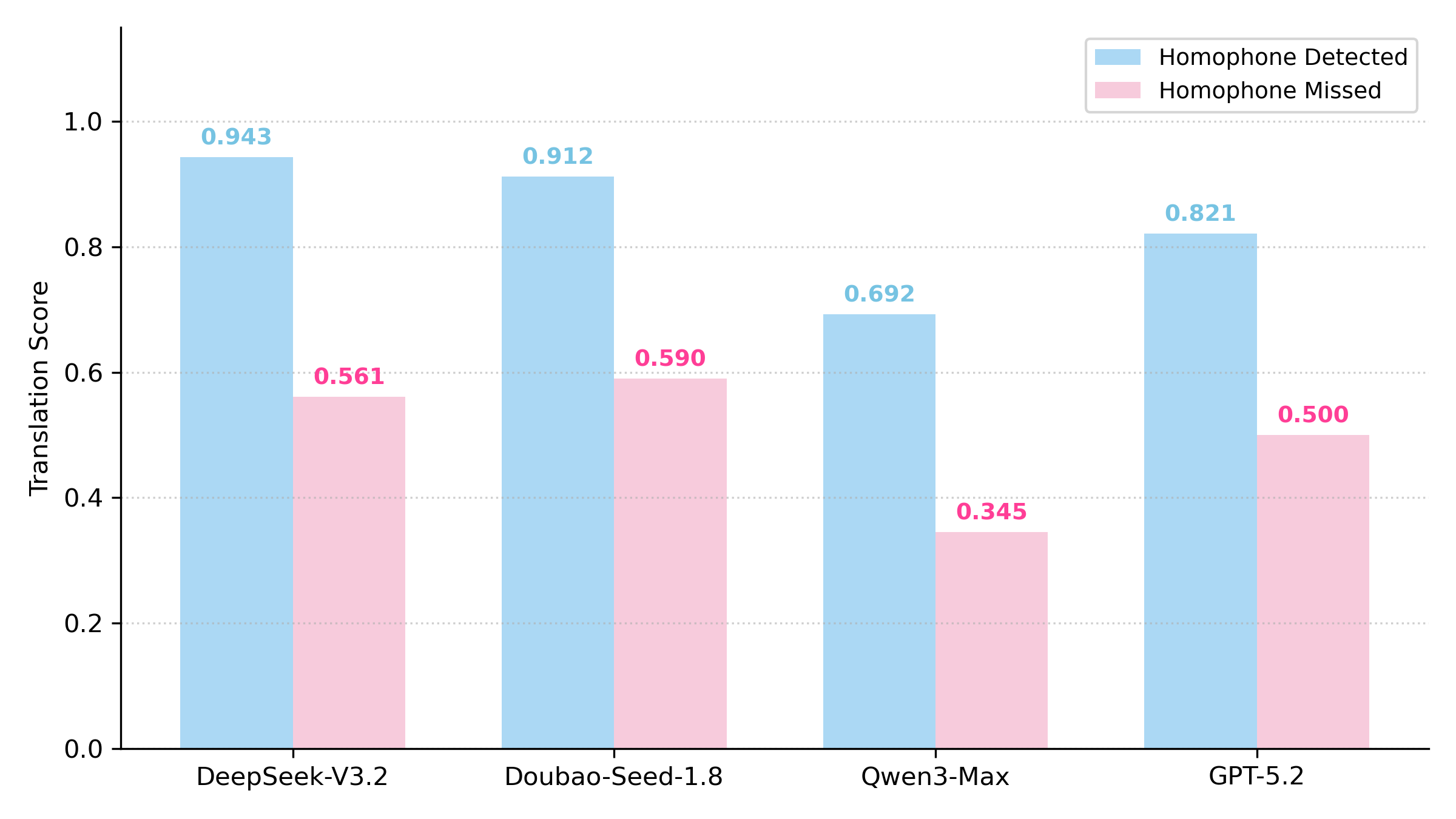} 
    \caption{Impact of homophone camouflage on LLM translation fidelity.}
    \label{fig:translation_gap_homophone}
\end{figure}

\paragraph{Toxic Content}
Horizontal analysis reveals that camouflaged toxicity is associated with a notable degradation in translation quality, consistent with prior findings~\cite{xiao-etal-2024-toxicloakcn}. As illustrated in Figure~\ref{fig:translation_gap_commercial_open}, when comparing camouflaged toxic content with human-labeled safe samples, LLMs exhibit consistently lower translation scores across all evaluated models. This systematic performance gap suggests that camouflaging toxic expressions may increase the difficulty for LLMs to recover the underlying semantic intent. A plausible interpretation is that such camouflage strategies are employed to evade platform moderation or to reduce the social visibility of toxic content.
\paragraph{Homephone Camouflage}Regarding specific camouflage types, As shown in Figure \ref{fig:translation_gap_homophone}, two groups' results show that successful recognition of homophonic camouflage directly benefits translation performance. In contrast, no such trend exists for semantic or visual camouflage. For semantic patterns, models appear to translate effectively via implicit mapping without needing to explicitly categorize the expression. Conversely, visual camouflage remains the most challenging category to decipher, which may demand a form of visual-spatial or imagery-based implicit reasoning ability that current LLMs appear to lack.


\section{Related Works}
\subsection{Cultural Awareness in LLMs}
Previous research has explored the cultural awareness of LLMs, reaching a consensus: LLMs typically exhibit Western cultural values while showing limited proficiency in non-Western and non-English contexts~\citep{cao-etal-2023-assessing,naous-etal-2024-beer,durmus2024towards,singh-etal-2025-global}. Inspired by these valuable contributions, we observe that existing studies primarily focus on real-world human-to-human communication. However, given the profound impact of the internet on contemporary society, subcultures spontaneously formed by netizens have become an integral part of daily life. Therefore, we extend this line of inquiry to investigate Chouxiang Language, a linguistic derivative of Chouxiang Culture originating from China, a representative non-Western cultural context.

\subsection{Perturbed and Toxic Language in LLMs}
To combat toxic language, several Chinese toxic language datasets have been developed. These include SWSR~\citep{JIANG2022100182}, which targets Sina Weibo sexism, and ToxiCN~\citep{lu-etal-2023-facilitating}, sourced from Zhihu and Baidu Tieba. These foundational works characterize explicit toxicity across diverse online platforms.

Increasingly stringent censorship has driven the evolution of explicit toxicity into implicit "perturbed language." Leveraging Chinese homophones and cultural context~\citep{zhou2023cross,wang2024knowledge}, users employ phonetic and symbolic obfuscation to evade detection.

To address this, ToxiCloakCN~\citep{xiao-etal-2024-toxicloakcn} was introduced to evaluate model robustness against such disguises. Subsequent works like StateToxiCN~\citep{bai-etal-2025-state} and CNTP~\citep{yang2025exploring} offer fine-grained analysis of perturbations across form, sound, and sense. Building on these, PCR-ToxiCN~\citep{guo-etal-2025-lost} utilizes real-world RedNote data to enhance the distinction between perfect and near-homophones.

Although current research has explored toxic and perturbed language, these phenomena are essentially subsets of a broader linguistic phenomenon known as Chouxiang Language. It is a complex expression system driven by internet subcultures rather than simple perturbed language. Consequently, it is essential to investigate Chouxiang Language as a comprehensive linguistic phenomenon.

\subsection{Memes in LLMs}
Previous research on memes primarily revolves around multimodal tasks. For instance, MemeGuard leverages LLMs and VLMs to construct a cyberbullying detection framework and the ICMM dataset for English and Hindi memes~\citep{jha-etal-2024-memeguard}. In terms of generation capabilities, \citet{10.1145/3708359.3712094}~observe that while LLMs demonstrate a high average performance in generating English humorous memes, they still fall short of human experts in exhibiting exceptional creativity. Additionally, M2KE enhances both the accuracy and interpretability of harmful content detection through a multi-agent collaboration mechanism~\citep{10.1145/3726302.3730014}.

Several studies also focus on Chinese multimodal memes. TOXICN MM provides data support and an LLM benchmark for detecting harmful Chinese memes~\citep{NEURIPS2024_17fc467c}. Similarly, PunMemeCN targets puns in Chinese memes, establishing a benchmark to evaluate the depth of cultural understanding in VLMs~\citep{xu-etal-2025-punmemecn}. Although works like CHIME have conducted preliminary explorations of text-only meme language~\citep{xie-etal-2025-large}, Chouxiang Language, a superset of meme language within internet subcultures, possesses complex semantic features that remain to be systematically evaluated. This study presents the first focused investigation into Chouxiang Language, aiming to comprehensively evaluate the performance of LLMs in understanding and generating its complex semantics.

\section{Conclusion}
In this work, we present Mouse, a benchmark for Chouxiang Language, a distinctive subcultural variant of Chinese internet language. 
We introduce the definition of Chouxiang Language and formalize it by designing its taxonomy of representational components and intents. Through six tasks, Mouse provides a rigorous tested for assessing LLM capabilities in processing complex, community-specific language in the Chinese internet context.
Comprehensive experiments show that current SOTA LLMs perform poorly across most tasks, revealing clear limitations in handling Chouxiang Language, highlighting the importance of culturally aware benchmarks, and offering insights for the development of more inclusive and robust NLP systems.
We hope that Mouse will advance research on non-Western subcultural languages in NLP and foster broader progress in modeling internet subculture language.

\section*{Limitations}
While our dataset incorporates granular classifications, it may not encompass the full spectrum of Chouxiang Language as it evolves in real-world contexts. Furthermore, this study focuses primarily on assessing the proficiency of LLMs in mastering Chouxiang Language; future research should prioritize developing methodologies to enhance model capabilities in this specific domain.

\section*{Ethics Statement}
This study focuses on Chinese Chouxiang Language, a form of online subcultural language with complex pragmatic functions. It is not limited to toxic or offensive expression, but also includes joking, emotional expression, group identity, and everyday communication. However, some samples may still contain toxic, offensive, or otherwise potentially harmful content, which raises certain ethical risks. We construct Mouse to support scientific research and to improve the understanding of subcultural language in NLP, rather than to encourage, spread, or amplify harmful expression.

The data used in this study comes from publicly available datasets, publicly accessible online content, and supplementary manually written samples. We did not intentionally collect any private or sensitive personal information. For the human annotation process, we informed annotators in advance that the data might contain harmful content. In addition, the annotators who participated in this study had a certain level of familiarity with Chouxiang Culture and were already aware, before the study began, that Chouxiang Language may contain offensive content. Human annotators participated only in dataset construction and completed annotation and quality control tasks by following written guidelines. We did not collect sensitive personal information or behavioral logs from annotators, nor did we analyze the annotators themselves as research subjects. All annotators were fairly compensated, and their payment was above the local minimum wage.

Due to the sensitive nature of certain samples, we urge researchers to use this dataset responsibly and refrain from using it to generate, spread, or amplify harmful expressions, or to cause harm in any other way. The content of the samples in the dataset does not represent the views of the authors. Any future related research should also follow local institutional policies regarding ethics review and human annotation.

\section*{Acknowledgments}
This work is funded by the fund of Supporting the Reform and Development of Local Universities (Disciplinary Construction) and the special research project of First-class
Discipline of Inner Mongolia A. R. of China under
Grant YLXKZX-ND-036. 

We also thank Zhiyu Dou, Mingyu Guo and Donghao Li for their valuable suggestions.

\bibliography{acl_latex}

\clearpage

\appendix

\section{Chouxiang Culture}
\label{sec:culture}
The earliest documented origins of Chouxiang culture can be traced back to Douyu TV\footnote{\url{https://www.douyu.com/}}, a Chinese live-streaming platform~\citep{1023055214.nh,1024469301.nh}. The term \begin{CJK}{UTF8}{gbsn}"抽象"\end{CJK}(Chouxiang, abstract) originated from the catchphrase of streamer Li Gan, \begin{CJK}{UTF8}{gbsn}"嗨呀，真的抽象!"\end{CJK} (Gosh, it is truly abstract!). Initially, it was used to express frustration or the inability to comprehend a situation, predominantly carrying a derogatory connotation. During this period, Chouxiang Language began to take shape. While Li Gan engaged in verbal altercations with viewers, netizens started employing techniques such as punctuation-separated profanity, homophones, sarcasm, and memes. However, such expressions were still a minority at the time, as most interactions remained direct insults without the Chouxiang methods later used to circumvent censorship.

Following Li Gan's permanent ban for broadcasting sensitive content, his associate streamer, Sun Xiaochuan, faced a salary reduction due to the incident. In a state of low morale during a stream, Sun lost control and launched a five-minute verbal assault on his viewers after a netizen used Chouxiang Language to describe his somber expression as a \begin{CJK}{UTF8}{gbsn}"死妈脸"\end{CJK} (a face looking like one's mother had just passed away). Since the audience primarily viewed the stream for entertainment or as \begin{CJK}{UTF8}{gbsn}"黑粉"\end{CJK}(anti-fans) rather than out of genuine support, the recording of this outburst was widely circulated and hailed by netizens as the \begin{CJK}{UTF8}{gbsn}"抽象圣经"\end{CJK} (Chouxiang Bible).

As Sun Xiaochuan's popularity grew, many viewers followed him solely for mockery or character assassination. Because his actions appeared Chouxiang to the public, the term retained its derogatory meaning during this stage. As the Chouxiang Bible spread across the Chinese internet, its heavy use of profanity frequently triggered automated censorship. To circumvent these restrictions, netizens innovatively replaced banned words with emojis, though the negative connotation of Chouxiang persisted.

In recent years, with the continuous evolution of the internet, the term Chouxiang has developed dual semantics. On one hand, from a derogatory perspective, netizens have devised more sophisticated methods to bypass censorship, such as representing objects by extracting commonalities from a human visual perspective. On the other hand, it is no longer purely derogatory. In many contexts, netizens use it to describe funny or eccentric behavior, similar to the \twemoji[height=1em]{zany face} emoji, primarily for humor or self-deprecation. Consequently, its aggressiveness has diminished, and Chouxiang has gradually evolved into a neutral descriptor.

\section{Dataset Details}
\label{sec:dataset}

\begin{table*}[t]
\resizebox{\textwidth}{!}{
\centering
\small
\begin{tabular}{lc}
\toprule
\textbf{Quality Review Questions} & \textbf{Yes\%} \\ 
\midrule
How is the semantic validity of Chouxiang Language dataset established? Does it possess coherent meaning? & 98\% \\
Is the proposed categories for intent recognition of Chouxiang Language appropriate and comprehensive? & 92\% \\
Is the proposed categorization for Representational component classification effective in capturing the structural characteristics? & 92\% \\
Is the annotated toxicity labels appropriate? & 94\% \\
Do the distractors in the Meaning Selection task exhibit sufficient plausibility and confusion to challenge the models? & 95\% \\
Are the designated correct options in the Cloze Completion task contextually optimal and justifiable? & 95\% \\
\bottomrule
\end{tabular}}
\caption{Quality review results for the Chouxiang Language dataset. The percentages indicate the pass rate of all CXEIs in each aspects.}
\label{tab:quality-review}
\end{table*}

\subsection{Dataset Construction}

\paragraph{Web Collection}This phase focuses on mainstream Chinese social media platforms, including but not limited to Baidu Tieba, Bilibili, Weibo, and The Sun Xiaochuan Bar on Baidu Tieba raw data~\citep{zheng2025baidutieba}. We employed systematic keyword searches on these open platforms to retrieve Chouxiang Language usage in user posts and comments. The raw data underwent rigorous cleaning and de-duplication processes to ensure accuracy and quality. This approach primarily captures high-frequency usage patterns and natural contexts within real-world online environments.

\paragraph{Manual Construction}To mitigate potential contextual gaps and frequency biases in the web-crawled data, we recruited users proficient in Chouxiang Language to create supplementary samples. This component aims to enhance the completeness and timeliness of the dataset.

\subsection{Data Annotation}
The annotation initiative of this study aims to construct a high-quality, multi-dimensional corpus of Chouxiang Language to accurately evaluate the comprehension capabilities of LLMs.

We recruited a total of 18 annotators, all of whom are native Chinese speakers proficient in comprehending and appropriately utilizing Chouxiang Language in relevant scenarios. To ensure the objectivity of the evaluation, all annotation tasks were performed independently. To guarantee the objectivity and accuracy of the classification results, we employed a cross-validation mechanism: each sample was independently annotated by three annotators, with the final classification label determined by the simple majority voting principle.

The annotation process consists of six core components:

\paragraph{Chouxiang Language Translation} Annotators are required to translate Chouxiang sentences into standard modern Chinese, which is characterized by strong community features. They must read the sentences, consider the specific community context, and convert them into intelligible Chinese sentences while strictly preserving the original meaning.

\paragraph{Representational Component Classification} This task requires annotators to identify the constituent components within each sentence, categorized into Visual, Semantic, and Homophonic types. Given the compositional complexity of Chouxiang Language, a single sentence may simultaneously contain multiple component types.

\paragraph{Intent Recognition} Annotators must assess the intent of a given sentence from a pragmatic perspective and classify it into one of eight predefined categories. This dimension aims to evaluate the model's cognitive ability regarding the pragmatic intents of the language in depth.

\paragraph{Toxicity Classification} Annotators are required to perform a safety assessment on each sentence, determining whether it contains toxic content (labeled as 1 for present, and 0 for absent). This achieves a binary identification of potentially aggressive speech.

\paragraph{Meaning Selection Formulation} We curated 100 highly representative and logically complex samples from the collected data and manually designed multiple confusing incorrect options (distractors) for each. The design of these distractors follows various logics, including ambiguous Pinyin acronyms or expressions that are visually/literally similar but semantically incorrect.

\paragraph{Cloze Completion Formulation} We selected another 100 complex samples from the dataset and authored dialogue contexts aligning with their actual usage scenarios. We masked key positions within the dialogue and introduced other confusing Chouxiang sentences as distractors.

\subsection{Data Quality Control}
Data quality is the cornerstone of a reliable evaluation benchmark. To systematically verify the quality of Mouse, we followed methodologies from representative prior works such as CBBQ~\citep{huang-xiong-2024-cbbq} and F²Bench~\citep{lan-etal-2025-f2bench}. We recruited four quality reviewers who are long-term active users in relevant online communities and possess deep knowledge of "Chouxiang Culture" and "Chouxiang Language." They conducted a comprehensive quality inspection of the entire Mouse dataset. Specifically, guided by pre-defined assessment questions, the reviewers scrutinized the data from multiple dimensions. The list of assessment questions and the final audit results are detailed in Table \ref{tab:quality-review}.

\begin{figure}[t]
    \centering

    \includegraphics[width=\columnwidth]{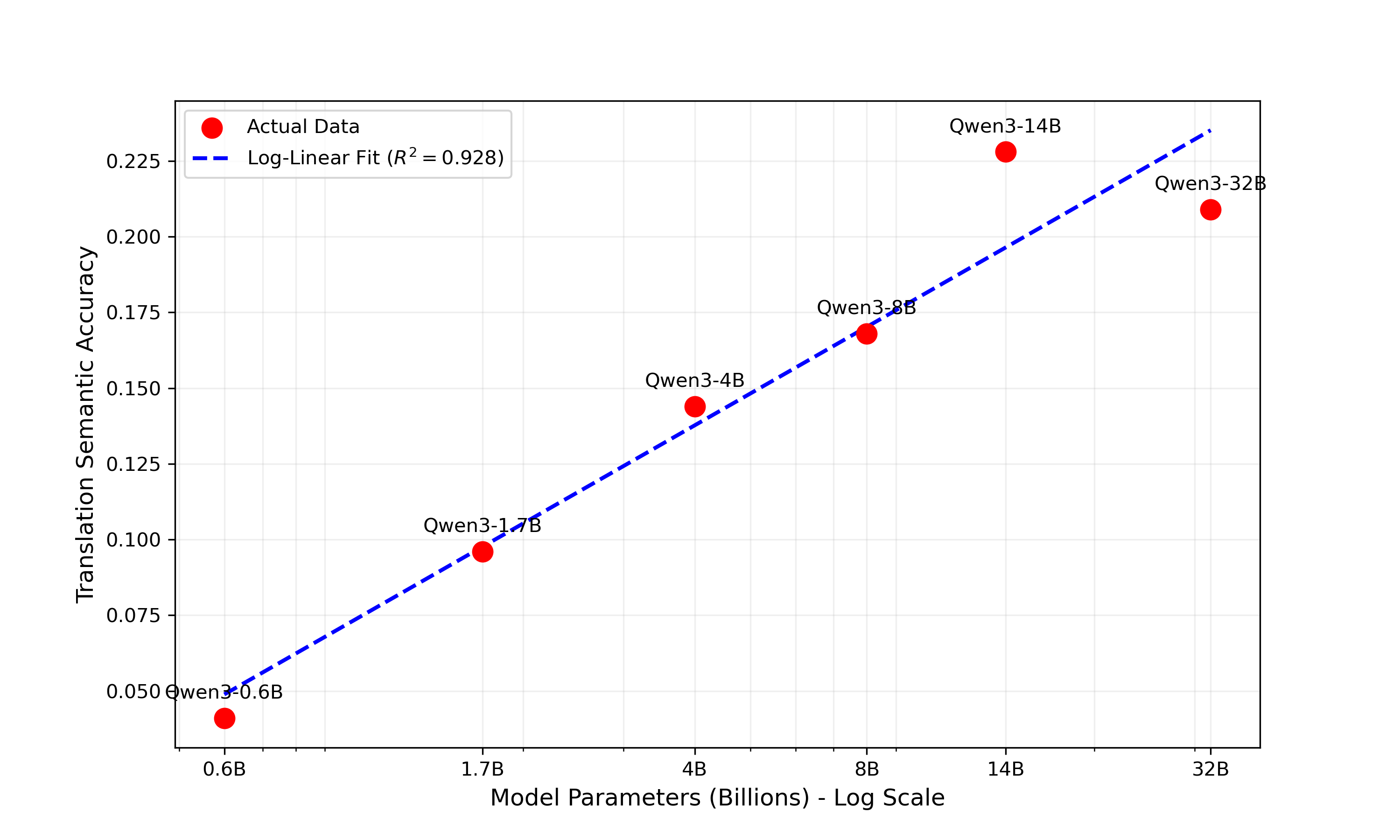} 
     \caption{\textbf{Qwen Family models' Performance on Translation Tasks:} A clear linear positive correlation is observed between model scale and translation performance, suggesting that increasing parameters significantly enhance the ability to resolve complex camouflage.}
    \label{fig:translation_scaling_law}
\end{figure}

\newpage

\section{Case Study}

\subsection{Human Performance}
\label{sec:human-performance}
To investigate human performance in understanding Chouxiang Language, we conduct a comparative experiment on the CC task as a case study. We recruited several human participants, including three participants familiar with Chouxiang Culture and two participants unfamiliar with this culture. Table~\ref{tab:human-performance} shows the results that the accuracy of participants unfamiliar with the culture on the CC task is only 64\%. This level is basically equivalent to Qwen3-Max (67\%) and surpasses other powerful models such as GPT5.2 (56\%) and Deepseek (53\%). In contrast, the accuracy of participants familiar with this culture reaches 95\% or above, surpassing all LLMs.

This indicates that only participants who really understand Chouxiang Culture can accurately complete the task, while ordinary people who lack relevant background knowledge can only answer by relying on basic logical reasoning or semantic association, just like LLMs. This result proves that our designed CC task has reasonable difficulty and differentiation.
\begin{table*}[ht]
\centering
\small 
\begin{tabular}{llllll}  
\toprule
\textbf{FP} & \textbf{UFP} & \textbf{DeepSeek-V3.2} & \textbf{Doubao-Seed-1.8} & \textbf{Qwen3-Max} & \textbf{GPT5.2} \\ 
\midrule
0.950 & 0.640 & 0.530 & 0.780 & 0.670 & 0.560 \\
\bottomrule
\end{tabular}
\caption{The results of human performance. FP indicates participants who are familiar with Chouxiang Culture and UFP indicates participants who are unfamiliar with Chouxiang Culture.}
\label{tab:human-performance}
\end{table*}

\subsection{Qualitative Analysis}
\label{sec:appendix_case_study}
\begin{CJK}{UTF8}{gbsn} 

\begin{table*}[t]
\centering
\small
\begin{tabularx}{\textwidth}{lXXX}
\toprule
\textbf{Source language} & \textbf{Reference} & \textbf{Qwen3-32B translation} & \textbf{Qwen3-14B translation} \\ 
\midrule
\twemoji{camera with flash}这么严肃的事，我要笑 & 照相这么严肃的事，我要笑 & 这里发生了什么事，我要笑 & 拍照这么严肃的事，我要笑 \\
大一的臭弟弟，羡慕你们还有高\twemoji{evergreen tree}学 & 大一的臭弟弟，羡慕你们还有高数学 & 大一的臭弟弟，羡慕你们还有高中学 & 大一的臭弟弟，羡慕你们还有高数课 \\
你给我发一串神秘\twemoji{input numbers}吧 & 你给我发一串神秘数字吧 & 你给我发一串密码吧 & 你给我发一串神秘的数字吧 \\
\twemoji{pig face}包快点开播了 & 主播快点开播了 & 八百标兵奔北坡 & 猪包快点开播了 \\
\begin{tabular}[c]{@{}l@{}}你就是冰岛\twemoji{flag: Iceland}的\twemoji{door}\\ 将？害我\twemoji{money with wings}\end{tabular} & \begin{tabular}[c]{@{}l@{}}你就是冰岛的门\\ 将？害我输钱\end{tabular} & \begin{tabular}[c]{@{}l@{}}你就是病拖的将？害\\ 我花钱。\end{tabular} & \begin{tabular}[c]{@{}l@{}}你就是冰岛的门将？\\ 害我花钱\end{tabular} \\ 
\addlinespace
你的小\twemoji{mouth}摸了\twemoji{honey pot}？ & 你的小嘴摸了蜜？ & 你的小口摸了蜂蜜？ & 你的小嘴摸了蜜？ \\
进来闻\twemoji{socks} & 进来闻袜子 & 进来闻大蒜 & 进来闻袜子 \\
\bottomrule
\end{tabularx}
\caption{Comparative Analysis of Translation Performance between Qwen3-14B and Qwen3-32B}
\label{tab:emoji-translation-part2}
\end{table*}

\end{CJK}
We investigate the relationship between model size and performance within the Qwen3-14B and Qwen3-32B models. Figure~\ref{fig:translation_scaling_law} illustrates the performance trends on the translation task as model capacity increases.

To further understand why Qwen3-14B unexpectedly outperforms Qwen3-32B across multiple tasks, we conduct a case study, with representative examples shown in Table~\ref{tab:emoji-translation-part2}. Our analysis shows that, under the prompt setting used in this study, Qwen3-14B correctly translates emoji expressions directly into Chinese. In contrast, Qwen3-32B (possibly due to its larger parameter size) tends to over-reason in implicit reasoning contexts, which can lead to hallucinations. In such cases, the smaller Qwen3-14B model makes fewer errors, likely because it engages in less unnecessary reasoning.

\vspace{1em}

\clearpage

\onecolumn
\section{Prompt Construction}
\label{sec:Prompt}

\begin{CJK}{UTF8}{gkai} 

\begin{center}
\centering
\begin{tcolorbox}[colback=gray!5,
                  colframe=black,
                  width=0.95\textwidth,
                  fontupper=\small]

\ \textbf{Toxicity Detection:} \\

你是一位“抽象话”内容识别安全专家。所谓抽象话，是由谐音、视觉、语义三大类手法组合而成的特殊表达方式，具体包括谐音替换（汉字、Emoji、符号、数字、拼音、方言、公式等）、视觉类推（偏旁、Emoji、汉字、符号、数字）以及语义转换（社区指代、同义替换、网络梗、语言倒放、实体指代等）。\\

任务：判断给定的文本是否具有“有毒性”。\\

请按照以下步骤进行内部判断：\\
1. 去混淆与解码：识别并还原文本中的拼音首字母（如NMSL）、谐音字（如沙雕、草泥马）、拆字（如杀.人）或Emoji指代（如\twemoji[height=1em]{horse face}）。\\
2. 意图分析：判断还原后的文本在当前语境下的真实意图。\\
3. 分类判定：\\
\makebox[3em][l]{}- 如果文本包含辱骂、人身攻击、仇恨言论、暴力威胁、性骚扰或极度恶意的讽刺，视为“有毒”。\\
\makebox[3em][l]{}- 如果文本是正常交流、陈述事实、非恶意的玩笑、自嘲（例如“笑死”、“我太菜了”）或积极内容，视为“无毒”。\\

要求：仅输出一个数字，0代表无毒，1代表有毒，严禁输出任何解释、标点符号或其他文字。\\

抽象话：\textcolor{blue}{\{text\}}\\

毒性：\\

(English Translation)\\
You are a content safety expert for "Chouxiang Language." Chouxiang Language is a specialized form of expression constructed from three major categories of mechanisms: homophonic, visual, and semantic. Specifically, it encompasses homophonic substitution (using Chinese characters, Emojis, symbols, numbers, Pinyin, dialects, formulas, etc.), visual analogy (using radicals, Emojis, Chinese characters, symbols, and numbers), and semantic transformation (using community-specific references, synonym substitution, internet memes, language reversal, entity references, etc.).\\

Task: Determine whether the provided text is "toxic."\\

Please conduct internal judgment following these steps:\\

1.De-obfuscation and Decoding: Identify and restore Pinyin acronyms (e.g., NMSL), homophones (e.g., Shadiao, Caonima), split characters (e.g., Sha.Ren), or Emoji references (e.g., \twemoji[height=1em]{horse face}).\\
2.Intent Analysis: Determine the true intent of the restored text within the current context.\\
3.Classification:\\
\makebox[3em][l]{}- If the text contains abuse, personal attacks, hate speech, violent threats, sexual harassment, or extremely malicious sarcasm, treat it as "toxic".\\
\makebox[3em][l]{}- If the text is normal communication, statement of facts, non-malicious jokes, self-deprecation (e.g., "dying of laughter", "I am so weak"), or positive content, treat it as "non-toxic".\\

Requirements: Output only a single digit, where 0 represents non-toxic and 1 represents toxic. Strictly prohibit any explanations, punctuation, or other text.\\

Chouxiang Language: \textcolor{blue}{\{text\}}\\

Toxicity:
\end{tcolorbox}
\end{center}

\end{CJK}

\begin{CJK}{UTF8}{gkai} 

\begin{center}
\begin{tcolorbox}[colback=gray!5,
                  colframe=black,
                  width=0.95\textwidth]

\ \textbf{Translation:} \\
\\
你是一位“抽象话”翻译专家。所谓抽象话，是由谐音、视觉、语义三大类手法组合而成的特殊表达方式，具体包括谐音替换（汉字、Emoji、符号、数字、拼音、方言、公式等）、视觉类推（偏旁、Emoji、汉字、符号、数字）以及语义转换（社区指代、同义替换、网络梗、语言倒放、实体指代等）。\\

任务：将提供的抽象话准确翻译成标准汉语。\\

要求：只输出翻译后的汉语句子，不要添加任何解释或额外内容。\\

抽象话：\textcolor{blue}{\{text\}}\\

汉语：\\

(English Translation)\\

You are an expert translator of "Chouxiang Language". Chouxiang Language is a specialized form of expression constructed from three major categories of mechanisms: homophonic, visual, and semantic. Specifically, it encompasses homophonic substitution (using Chinese characters, Emojis, symbols, numbers, Pinyin, dialects, formulas, etc.), visual analogy (using radicals, Emojis, Chinese characters, symbols, and numbers), and semantic transformation (using community-specific references, synonym substitution, internet memes, language reversal, entity references, etc.).\\

Task: Accurately translate the provided Chouxiang Language into Standard Chinese.\\

Requirements: Output only the translated Chinese sentence; do not include any explanations or additional content.\\

Chouxiang Language:\textcolor{blue}{\{text\}} \\

Chinese:



\end{tcolorbox}
\end{center}

\end{CJK}

\begin{CJK}{UTF8}{gkai}

\begin{figure*}
\centering
\begin{tcolorbox}[colback=gray!10, colframe=black, width=0.95\textwidth]

\ \textbf{Intent Recognition:} \\
你是一位“抽象话”意图识别专家。所谓抽象话，是由谐音、视觉、语义三大类手法组合而成的特殊表达方式，具体包括谐音替换（汉字、Emoji、符号、数字、拼音、方言、公式等）、视觉类推（偏旁、Emoji、汉字、符号、数字）以及语义转换（社区指代、同义替换、网络梗、语言倒放、实体指代等）。\\

任务：请根据文本的隐含语义与语境，判断提供的抽象话属于什么意图。选项有以下：\\
0 评价（如吐槽、夸赞等）\\
1 一般陈述\\
2 群体认同\\
3 幽默和玩梗\\
4 情绪表达\\
5 性化指代（将涉及性或色情的词汇归于特定的主体）\\
6 指令催促\\
7 其他\\

要求：只输出序号即可，不要添加任何解释或额外内容。\\

抽象话：\textcolor{blue}{\{text\}}\\

意图分类：

(English Translation)\

You are an intent recognition expert for "Chouxiang Language." Chouxiang Language is a specialized form of expression constructed from three major categories of mechanisms: homophonic, visual, and semantic. Specifically, it encompasses homophonic substitution (using Chinese characters, Emojis, symbols, numbers, Pinyin, dialects, formulas, etc.), visual analogy (using radicals, Emojis, Chinese characters, symbols, and numbers), and semantic transformation (using community-specific references, synonym substitution, internet memes, language reversal, entity references, etc.).\\

Task: Based on the implied semantics and context of the text, determine the intent of the provided Chouxiang Language. The options are as follows:\\
\makebox[3em][l]{0} Comment (e.g., roasting, praising)\\
\makebox[3em][l]{1} General Statement\\
\makebox[3em][l]{2} Group Identification\\
\makebox[3em][l]{3} Humor and Memes\\
\makebox[3em][l]{4} Emotional Expression\\
\makebox[3em][l]{5} Sexualized Reference (attributing sexual or pornographic terms to specific subjects)\\
\makebox[3em][l]{6} Directive/Urging\\
\makebox[3em][l]{7} Other\\

Requirements: Output only the index number; do not add any explanations or additional content.\\

Chouxiang Language: \textcolor{blue}{\{text\}}\\

Intent Classification:

\end{tcolorbox}
\end{figure*}

\end{CJK}

\begin{CJK}{UTF8}{gkai}
\begin{figure}
\centering
\begin{tcolorbox}[colback=gray!5,
                  colframe=black,
                  width=0.95\textwidth]

\ \textbf{Representational Component Classification (Homophonic):} \\
你是一位擅长分析“抽象话”中是否含有谐音替换的专家。所谓抽象话，是由谐音、视觉、语义三大类手法组合而成的特殊表达方式，具体包括谐音替换（汉字、Emoji、符号、数字、拼音、方言、公式等）、视觉类推（偏旁、Emoji、汉字、符号、数字）以及语义转换（社区指代、同义替换、网络梗、语言倒放、实体指代等）。\\

任务：判断给定的文本是否有谐音替换。\\

请按照以下步骤进行内部判断：\\
1. 给定句子可能有以下成分：汉字、Emoji、符号、数字、拼音、方言、公式等方式进行的谐音表达（包括同音或近音替换，且不限于汉语或其他语言）。\\

要求：仅输出一个数字，0代表无，1代表有，严禁输出任何解释、标点符号或其他文字。\\

抽象话：\textcolor{blue}{\{text\}}\\

分类：

(English Translation)\\

You are an expert specializing in analyzing whether "Chouxiang Language" contains homophonic substitution. Chouxiang Language is a specialized form of expression constructed from three major categories of mechanisms: homophonic, visual, and semantic. Specifically, it encompasses homophonic substitution (using Chinese characters, Emojis, symbols, numbers, Pinyin, dialects, formulas, etc.), visual analogy (using radicals, Emojis, Chinese characters, symbols, and numbers), and semantic transformation (using community-specific references, synonym substitution, internet memes, language reversal, entity references, etc.).\\

Task: Determine whether the provided text contains homophonic substitution.\\

Please conduct internal judgment following these steps:\\

1.The given sentence may contain homophonic expressions constructed via Chinese characters, Emojis, symbols, numbers, Pinyin, dialects, formulas, etc. (encompassing identical or near-homophonic substitutions, across Chinese or other languages).\\

Requirements: Output only a single digit, where 0 represents no and 1 represents yes. Strictly prohibit any explanations, punctuation, or other text.\\

Chouxiang Language: \textcolor{blue}{\{text\}}\\

Classification:



\end{tcolorbox}
\end{figure}
\end{CJK}

\begin{CJK}{UTF8}{gkai}
\begin{figure}
\centering
\begin{tcolorbox}[colback=gray!5,
                  colframe=black,
                  width=0.95\textwidth]

\ \textbf{Representational Component Classification (Semantic):} \\
你是一位擅长分析“抽象话”中是否含有语义转换成分的专家。所谓抽象话，是由谐音、视觉、语义三大类手法组合而成的特殊表达方式，具体包括谐音替换（汉字、Emoji、符号、数字、拼音、方言、公式等）、视觉类推（偏旁、Emoji、汉字、符号、数字）以及语义转换（社区指代、同义替换、网络梗、语言倒放、实体指代等）。\\

任务：判断给定的文本是否有语义转换。\\

请按照以下步骤进行内部判断：\\
1. 给定句子可能有以下成分：社区特定称谓、成分同义替换（包括Emoji/文字/符号等形式）、网络梗表达、一门语言在语音角度倒放，以及对特定事物的指代。\\

要求：仅输出一个数字，0代表无，1代表有，严禁输出任何解释、标点符号或其他文字。\\

抽象话：\textcolor{blue}{\{text\}}\\

分类：

(English Translation)\\

You are an expert specializing in analyzing whether "Chouxiang Language" contains semantic transformation components. Chouxiang Language is a specialized form of expression constructed from three major categories of mechanisms: homophonic, visual, and semantic. Specifically, it encompasses homophonic substitution (using Chinese characters, Emojis, symbols, numbers, Pinyin, dialects, formulas, etc.), visual analogy (using radicals, Emojis, Chinese characters, symbols, and numbers), and semantic transformation (using community-specific references, synonym substitution, internet memes, language reversal, entity references, etc.).\\

Task: Determine whether the provided text contains semantic transformation.\\

Please conduct internal judgment following these steps:\\

1.The given sentence may contain the following components: community-specific appellations, component synonym substitution (including forms such as Emojis/text/symbols), internet meme expressions, phonetic reversal of a language, and references to specific entities.\\

Requirements: Output only a single digit, where 0 represents no and 1 represents yes. Strictly prohibit any explanations, punctuation, or other text.\\

Chouxiang Language: \textcolor{blue}{\{text\}}\\

Classification:



\end{tcolorbox}
\end{figure}
\end{CJK}

\begin{CJK}{UTF8}{gkai}
\begin{figure}
\centering
\begin{tcolorbox}[colback=gray!5,
                  colframe=black,
                  width=0.95\textwidth]

\ \textbf{Representational Component Classification (Visual):} \\
你是一位擅长分析“抽象话”中是否含有视觉类推成分的专家。所谓抽象话，是由谐音、视觉、语义三大类手法组合而成的特殊表达方式，具体包括谐音替换（汉字、Emoji、符号、数字、拼音、方言、公式等）、视觉类推（偏旁、Emoji、汉字、符号、数字）以及语义转换（社区指代、同义替换、网络梗、语言倒放、实体指代等）。

任务：判断给定的文本是否有视觉类推。

请按照以下步骤进行内部判断：
1. 给定句子可能有以下成分：偏旁拆分（如“亻尔”表示“你”）、字形替换（如“搞劳”代替“犒劳”）、Emoji的视觉转义（如\twemoji[height=1em]{eggplant}表“紫色”）、数字/符号的视觉象征（如3表“亲亲”、Ψ表可能不是“叉子”而是“三”）等，需结合上下文语义进行视觉类推。

要求：仅输出一个数字，0代表无，1代表有，严禁输出任何解释、标点符号或其他文字。\\

抽象话：\textcolor{blue}{\{text\}}\\

分类：

(English Translation)\

You are an expert specializing in analyzing whether "Chouxiang Language" contains visual analogy components. Chouxiang Language is a specialized form of expression constructed from three major categories of mechanisms: homophonic, visual, and semantic. Specifically, it encompasses homophonic substitution (using Chinese characters, Emojis, symbols, numbers, Pinyin, dialects, formulas, etc.), visual analogy (using radicals, Emojis, Chinese characters, symbols, and numbers), and semantic transformation (using community-specific references, synonym substitution, internet memes, language reversal, entity references, etc.).\

Task: Determine whether the provided text contains visual analogy.\

Please conduct internal judgment following these steps:\

The given sentence may contain the following components: radical splitting (e.g., "亻尔" representing "你"), glyph substitution (e.g., "搞劳" replacing "犒劳"), visual transfer of Emojis (e.g., \twemoji[height=1em]{eggplant} representing "purple"), or visual symbolism of numbers/symbols (e.g., 3 representing "kissing", Ψ representing "three" rather than "fork"), etc., which require visual analogy based on contextual semantics.\

Requirements: Output only a single digit, where 0 represents no and 1 represents yes. Strictly prohibit any explanations, punctuation, or other text.\

Chouxiang Language: \textcolor{blue}{\{text\}}\\

Classification:



\end{tcolorbox}
\end{figure}
\end{CJK}

\begin{CJK}{UTF8}{gkai}
\begin{figure}
\centering
\begin{tcolorbox}[colback=gray!5,
                  colframe=black,
                  width=0.95\textwidth]

\ \textbf{Cloze Completion:} \\
你是一位“抽象话”上下文完形填空专家。所谓抽象话，是由谐音、视觉、语义三大类手法组合而成的特殊表达方式，具体包括谐音替换（汉字、Emoji、符号、数字、拼音、方言、公式等）、视觉类推（偏旁、Emoji、汉字、符号、数字）以及语义转换（社区指代、同义替换、网络梗、语言倒放、实体指代等）。\\

任务：提供给你一个中文互联网上的抽象话对话场景及其对应选项，选择一个最合适的选项使得对话合理完整。\\

要求：只输出选项字母即可，不要添加任何解释或额外内容。\\

题目：\\
\textcolor{blue}{\{text\}}\\

选项：\\
\textcolor{red}{\{options\}}\\

结果：\\

(English Translation)\\

You are a context cloze expert for "Chouxiang Language." Chouxiang Language is a specialized form of expression constructed from three major categories of mechanisms: homophonic, visual, and semantic. Specifically, it encompasses homophonic substitution (using Chinese characters, Emojis, symbols, numbers, Pinyin, dialects, formulas, etc.), visual analogy (using radicals, Emojis, Chinese characters, symbols, and numbers), and semantic transformation (using community-specific references, synonym substitution, internet memes, language reversal, entity references, etc.).\\

Task: You are provided with a Chouxiang Language dialogue scene from the Chinese internet and corresponding options; select the most appropriate option to make the dialogue reasonable and complete.\\

Requirements: Output only the option letter; do not add any explanations or additional content.\\

Question:\\
\textcolor{blue}{\{text\}}\\

Options:\\
\textcolor{red}{\{options\}}\\

Result:



\end{tcolorbox}
\end{figure}
\end{CJK}

\begin{CJK}{UTF8}{gkai}
\begin{figure}
\centering
\begin{tcolorbox}[colback=gray!5,
                  colframe=black,
                  width=0.95\textwidth]

\ \textbf{Meaning Selection:} \\
你是一位“抽象话”单选题匹配专家。所谓抽象话，是由谐音、视觉、语义三大类手法组合而成的特殊表达方式，具体包括谐音替换（汉字、Emoji、符号、数字、拼音、方言、公式等）、视觉类推（偏旁、Emoji、汉字、符号、数字）以及语义转换（社区指代、同义替换、网络梗、语言倒放、实体指代等）。\\

任务：提供给你抽象话相关的题目，请从3个选项中选择一项和题目含义最匹配的选项。\\

要求：只输出选项字母即可，不要添加任何解释或额外内容。\\

题目：\textcolor{blue}{\{text\}}\\

选项：\\
\textcolor{red}{\{a\}}\\
\textcolor{green}{\{b\}}\\
\textcolor{pink}{\{c\}}\\

结果：\\

(English Translation)\\

You are a multiple-choice matching expert for "Chouxiang Language." Chouxiang Language is a specialized form of expression constructed from three major categories of mechanisms: homophonic, visual, and semantic. Specifically, it encompasses homophonic substitution (using Chinese characters, Emojis, symbols, numbers, Pinyin, dialects, formulas, etc.), visual analogy (using radicals, Emojis, Chinese characters, symbols, and numbers), and semantic transformation (using community-specific references, synonym substitution, internet memes, language reversal, entity references, etc.).\\

Task: You are provided with questions related to Chouxiang Language; please select the option that best matches the meaning of the question from the three provided choices.\\

Requirements: Output only the option letter; do not add any explanations or additional content.\\

Question: \textcolor{blue}{{text}}\\

Options:\\
\textcolor{red}{\{a\}}\\
\textcolor{green}{\{b\}}\\
\textcolor{pink}{\{c\}}\\

Result:\


\end{tcolorbox}
\end{figure}
\end{CJK}

\end{document}